\documentclass[10pt,twocolumn,letterpaper]{article}

\usepackage{cvpr}
\usepackage{times}
\usepackage{epsfig}
\usepackage{graphicx}
\usepackage{amsmath}
\usepackage{amssymb}
\usepackage{tabulary,multirow,xspace,xcolor}
\usepackage{setspace}
\usepackage{caption}

\usepackage[pagebackref=true,breaklinks=true,letterpaper=true,colorlinks,bookmarks=false]{hyperref}

\cvprfinalcopy % *** Uncomment this line for the final submission

\def\cvprPaperID{****} % *** Enter the CVPR Paper ID here

\newif\ifshowcomment
\newcommand{\xxcom}[2]{\ifshowcomment\textcolor{#1}{#2}\fi}
\newcommand{\gscom}[1]{\xxcom{red}{greg: #1}} % greg's comment
\newcommand{\rtcom}[1]{\xxcom{purple}{rt: #1}} % rt

\ifcvprfinal\fi
\begin{document}

%%%%%%%%% TITLE
\title{Controlling Length in Image Captioning}

\author{Ruotian Luo\\
TTI-Chicago\\
{\tt\small rluo@ttic.edu}
\and
Greg Shakhnarovich\\
TTI-Chicago\\
{\tt\small greg@ttic.edu}
}

\maketitle
%\thispagestyle{empty}

%%%%%%%%% ABSTRACT
\begin{abstract}
    We develop and evaluate captioning models that allow control of
    caption length. Our models can leverage this control to generate captions of different style and descriptiveness.
\end{abstract}

%%%%%%%%% BODY TEXT
\section{Introduction}

Most existing captioning models learn an autoregressive model, either
LSTM or transformer, in which explicit control of generation process
is difficult. In particular, the length of the caption is determined
only after the End Of Sentence ($<$eos$>$) is generated. It is hard to
know and control the length beforehand. However, length can be an
important property of a caption. A model that allows control of output
length would provide more options for the end users of the captioning
model. By controling the length, we can
influence the style and descriptiveness of the caption: short, simple
captions vs. longer, more complex and detailed descriptions for the
same image. 

Previous work includes captioning models that allow control for other aspects. \cite{cornia2019show} controls the caption by inputting a different set of image regions. \cite{deshpande2019fast} can generate a caption controlled by assigning POS tags.
% One simple but interesting component to control is length where has been largely neglected in captioning field. While there are some length control methods like length normalization\cite{hossain2019comprehensive} borrowed from NLP task like NMT, these methods can only loosly control the length.
Length control has been studied in abstract
summarization~\cite{kikuchi2016controlling,fan2017controllable,takase2019positional},
but to our knowledge not in the context of image capitoning.

To control the length of the generated caption, we build the model
borrowing existing ideas from summarization work, by injecting length
information into the model. To generate captions without an explicit
length specification, we add a \emph{length prediction module} that
can predict the optimal length for the input image at hand.
We show that the length models can successfully generate catpion
ranging from 7 up to 28 words long. We also show that length models
perform better than non-controled model (with some special decoding
methods)\gscom{the last phrase is vague at this point} when asked to generate long captions.

\section{Models}
% We base our method on existing LSTM based captioning model. The method is also applicable to Transformer-based models, but we limit our analysis on LSTM.

We consider repurposing existing methods in summarization for captioning.
% We modify these model by introducing the length prediciton model.
In general, the length is treated as an intermediate \rtcom{latetnt variable is not correct, it's not latent}
variable: $P(c|I) =
P(c|I,l)*P(l|I)$. $c$, $I$ and $l$ are caption, image and length,
respectively. We introduce how we build $P(c|I,l)$ and $P(l|I)$ as
follows. Note that, the following methods can be used in conjunctions
with any standard captioning model.
% \gscom{is that what you meand by ``applicable''?}

\subsection{LenEmb\cite{kikuchi2016controlling}}
We take LenEmb from \cite{kikuchi2016controlling} and make some small
change according to \cite{takase2019positional}. Given desired length
$T$ and current time step $t$, we embed the remaining length $T-t$
into a vector that is the same size as the word embedding. Then the
word embedding of the previous word $w_{t-1}$, added with the length
embedding (rather than concatenated, as in \cite{kikuchi2016controlling}), is fed as the input to the rest of the LSTM model.
\begin{align}
x_t = E_w[w_{t-1}] + E_l[T-t];\ \ P(w_t) = LSTM(x_t, h_{t-1}) \nonumber
\end{align}

\noindent\textbf{Learning to predict length} We add a length prediction
module to predict the length given the image features (in this case
the averaging region features) while no desired length is
provided. We treat it as a classification task and train it
with the reference caption length.

% To inject the remaining length information into the model, we borrow the design from \cite{kikuchi2016controlling} and \cite{takase2019positional}. At each time step, we have an additional remaining length embedding. In \cite{kikuchi2016controlling}, the embedding is learned and is concatenated (\rtcom{verify!}) with the word embedding, while \cite{takase2019positional} use a fixed positional embedding and apply addition with the word embedding. For simplicity, we combine both design, use a learned embedding and add it with the word embedding. Empirically, we find concatenation and addition achieve similar performance.

% \subsection{LenInit\cite{kikuchi2016controlling}}
% The memory cell is initialized to be length times a learnable vector.

\subsection{Marker\cite{fan2017controllable}}
We also implement Marker model from \cite{fan2017controllable}. The desired length is fed as a special token at the beginning of generation as the ``first word''. At training time, the model needs to learn to predict the length at the first step the same way as other words (no extra length predictor needed). At test time, the length token is sampled in the same way as other words if no desired length specified.

% \subsection{LenEmb+marker}
% \rtcom{marker doesn't predict, marker uses fixed length}

% \subsection{Training with self critical}
% With the finding in my diversity paper, self-critical training would lead to peaky distribution. To alleviate this for length control model, we try several ways during training:
% 1. fix the length prediction MLP. This allows the model to maintain the length distribution. (Potentially the reuslts show that the length distribution on validation on test set fits much better compared to xx)
% 2. Uniform distribution. For model to explore its best output for each length.

% To let the length control obey the length. Two ways. 1, use constrained.
% Note that the logprobs used should be the modified ones.
% 2. Put it in the reward.

\section{Experiments}
We use COCO~\cite{lin2014microsoft} to evaluate our models. For train/val/test split we follows \cite{karpathy2015deep}.
The base captioning model is Att2in~\cite{rennie2017self}. The images features are bottom-up features\cite{anderson2018bottom}.
For evaluation, we use BLEU~\cite{papineni2002bleu}, METEOR~\cite{denkowski2014meteor}, ROUGE~\cite{lin2004rouge}, CIDEr~\cite{vedantam2015cider}, SPICE~\cite{anderson2016spice} and bad ending rate~\cite{guo2018improving,self-critical-code}.
We train the models with cross-entropy loss. Unless specified
otherwise, decoding is beam search with beam size 5, and evaluation on Karpthy test set.
% \subsection{Golden length}
% We can also use normal model with constrained decoding to compare.

% Unlike most abstraction dataset where only one reference is provided, commonly used captioning dataset has more than one references. To choose a golden length during test time. We use several type, one, median of the length, max of the length, min of the length.

% Transformer, At test time, three lengths are generated given the ground truth length, 10,23,50?

% Should we use block trigram (Fan did that)

% \subsection{Short/Medium/Long}

% fixrng needs bigger beam size

\subsection{Generation with predicted lengths}
For fair comparison on general image captioning task, we predict the length and generate the caption conditioned on the predicted length for length models. Results in Table \ref{tab:pred_perf} show that the length models are comparable to the base model.

\begin{table}[htbp]
\footnotesize
  \centering
    \begin{tabular}{lccccccc}
          & B4 & R & M & C & S & BER & LenMSE\\
    \hline
    Att2in & 35.9 & 56.1  & 27.1  & 110.6 & 20.0  & 0.1 & N/A\\
    \hline
    LenEmb & 34.9 & 56.2  & 27.0  & 110.0 & 20.0  & 0.0 & 0\\
    Marker & 35.2 & 56.2  & 26.9  & 109.8 & 19.9  & 0.0 & 0\\
    \hline
    \end{tabular}%
  \caption{\small Performance on COCO Karpathy test set. B4=BLEU4, R=ROUGE, M=METEOR, C=CIDEr, S=SPICE, BER=bad ending rate, LenMSE=length mean square error. Numbers all in percentage(\%) except LenMSE.}
  \label{tab:pred_perf}%
\end{table}%

\begin{figure}[t]
\centering
\begin{minipage}[b]{.6\linewidth}
\includegraphics[width=\linewidth]{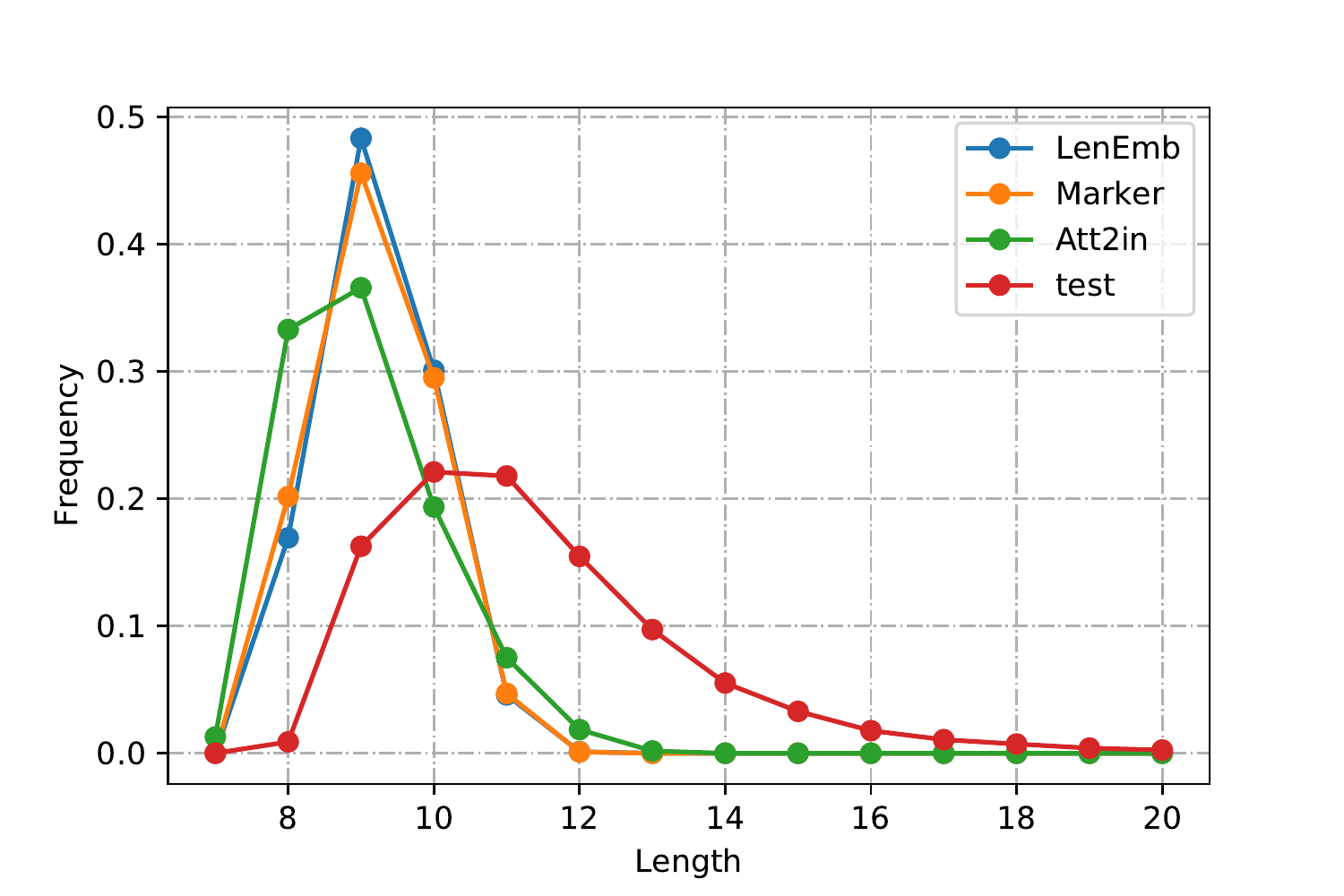}
\end{minipage}
\begin{minipage}[b]{.38\linewidth}
\captionof{figure}{\small Length distribution of captions generated by different models. For length models, the length is obtained within the beam search process as a special token.}
\label{fig:length_dist}
\end{minipage}
\end{figure}

\noindent\textbf{Length distribution(Fig.\ref{fig:length_dist})}
While the scores are close, the length distribution is quite different. Length models tend to generate longer captions than normal auto-regressive models. However neither is close to the real caption length distribution(``test'' in the figure).

% We can sample the length first and then apply beam search fixing the length. This gives us worse CIDEr(1.00) but close the the real distribution (however, the length predicted seems to be a little shorter.)

% Sample the length then beam search.

\subsection{Generation with controlled lengths}
For baseline model, we use the method fixLen in \cite{kikuchi2016controlling} where probability manipulation are used to avoid generating eos token until desired length.

The original CIDEr-D promotes short and generic captions 
%that cover the most common things among the reference captions
because it computes average similarity between the generated and the references. We report a modified CIDEr (mCIDEr): 1) removing length penalty term in the CIDEr-D; 2) combining the ngram counts from all the reference captions to compute similarity~\cite{coco-caption-code}.

\noindent\textbf{Fluency} The high bad ending rate for Att2in indicates it can't generate fluent sentences;
when increasing beam size, the bad ending rate becomes lower. For length models,
Marker performs well when length is less than 20 but collapse after, while LenEmb performs consistently well.

\noindent\textbf{Accuracy} The length models perform better than base model when length$<10$. The base model performs better between 10-16 which are the most common lengths in the dataset. For larger length, the LenEmb performs the best on both mCIDEr and SPICE, incidcating it's covering more information in the reference captions.

% CIDEr drops as length increase while SPICE increases.

\begin{figure}[t]
\centering
\includegraphics[width=0.45\linewidth]{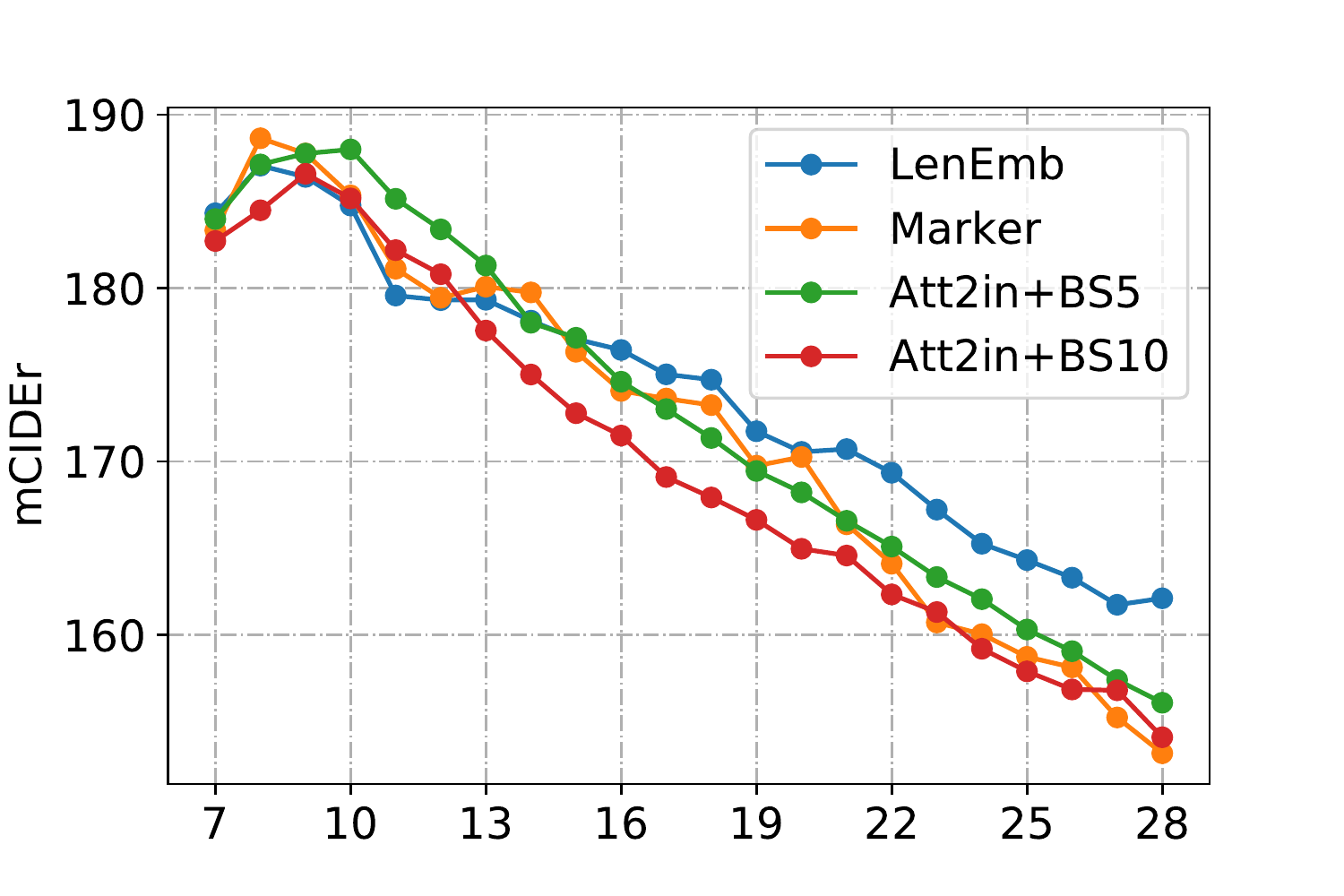}
\includegraphics[width=0.45\linewidth]{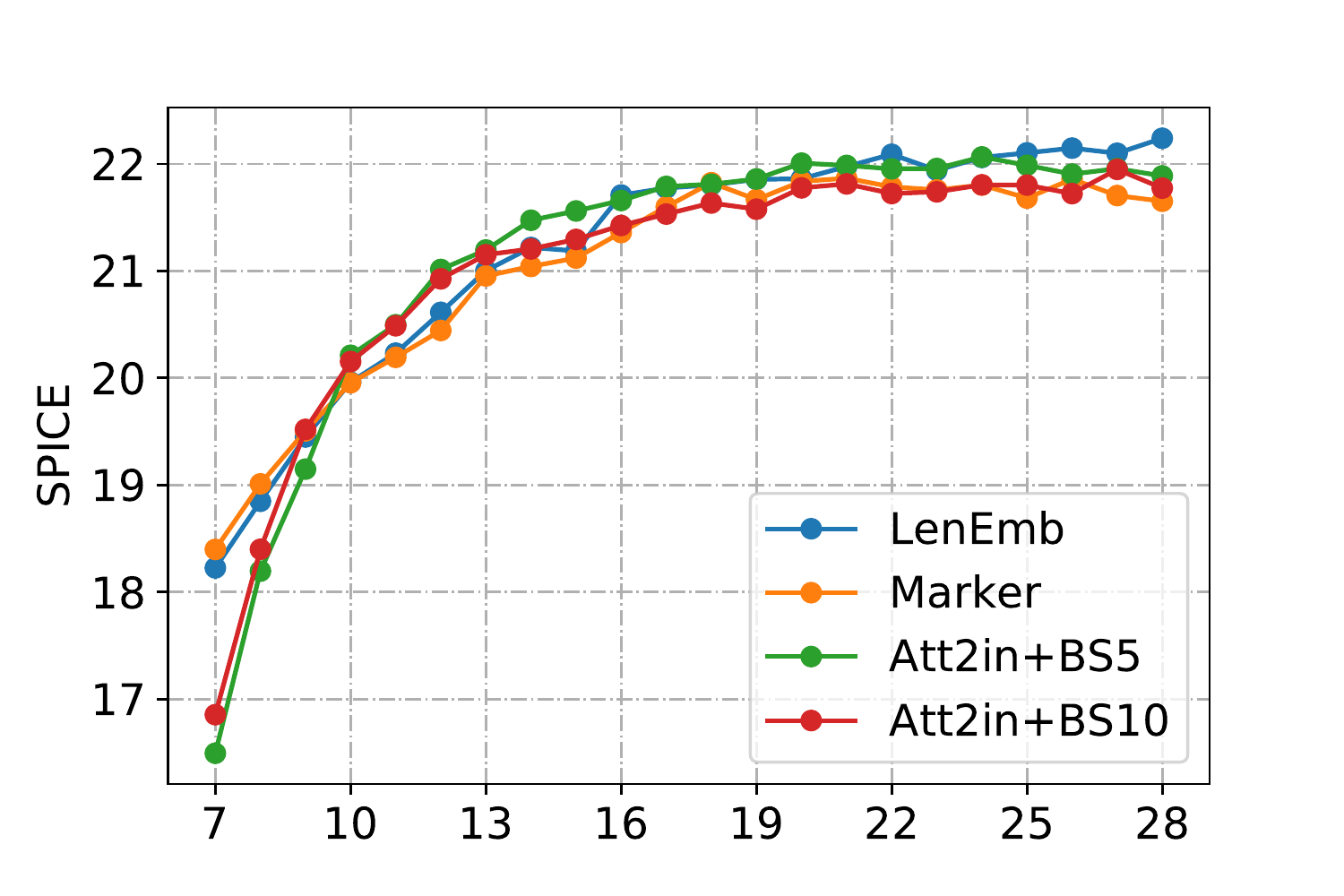}
\includegraphics[width=0.45\linewidth]{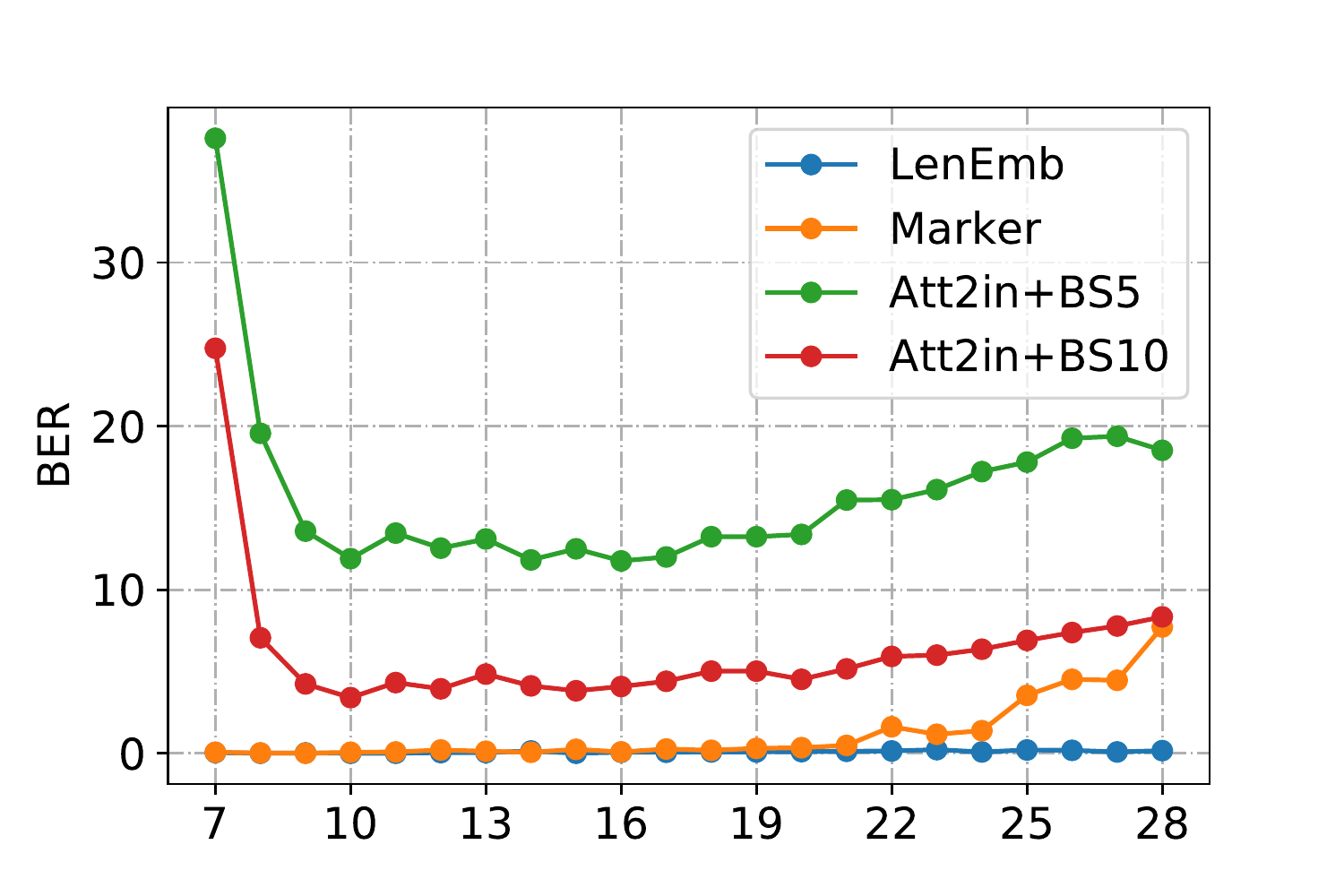}
\includegraphics[width=0.45\linewidth]{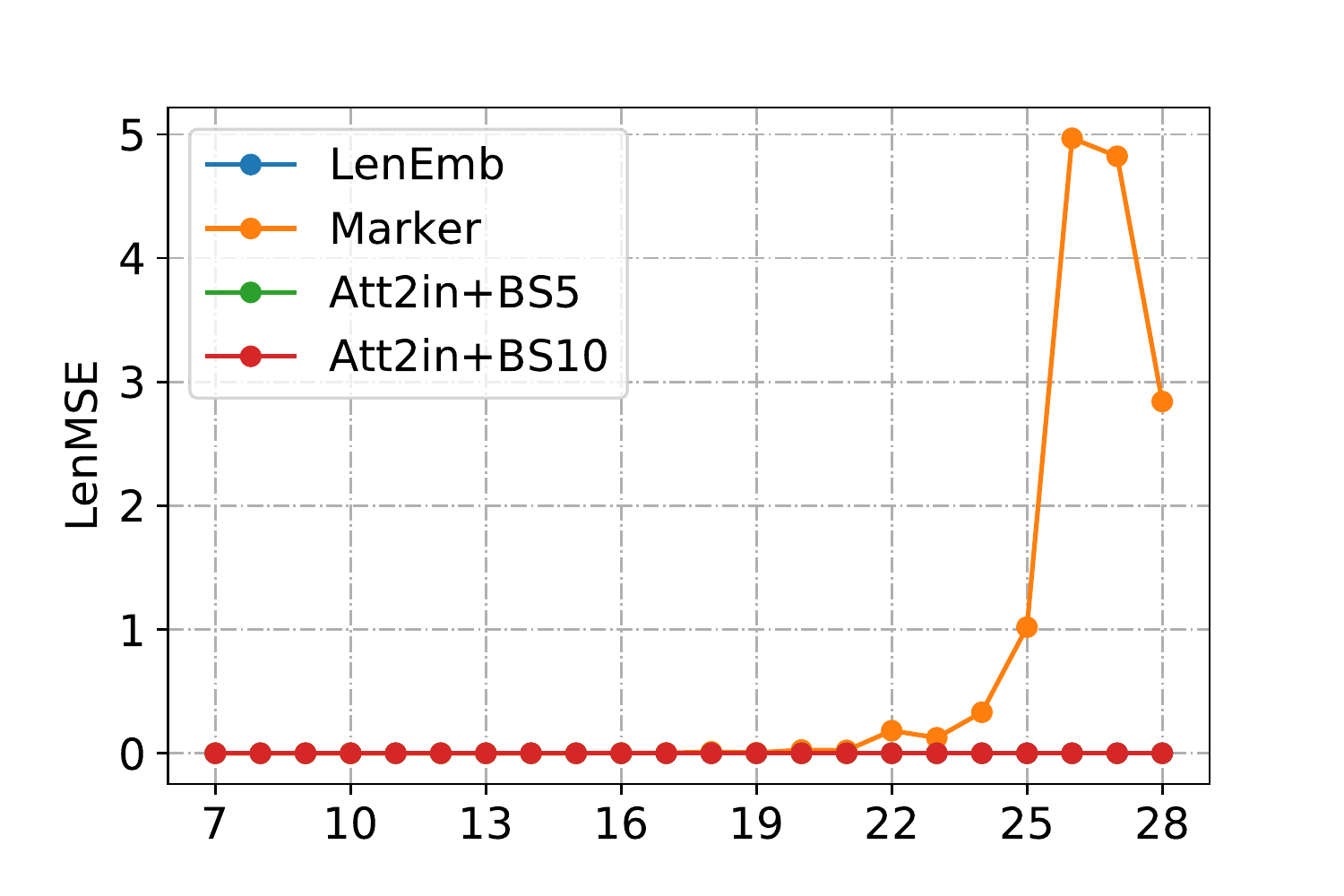}
\caption{\small The performance of models with different desired length. Att2in+BSx is Att2in+fixLen with beam size x.}
\label{fig:scores}
\end{figure}

\noindent\textbf{Controllability} We use the mean square error between the desired length and the actual length (LenMSE) to evaluate the controllability. When using predicted length, the length models perfectly achieve the predicted length (in Table \ref{tab:pred_perf}). When desired length is fed, Fig. \ref{fig:scores} shows that LenEmb can perfectly obey the length while Marker fails for long captions probabily due to poor long-term dependency.

\begin{figure}[!th]
{\footnotesize
  \centering
    \begin{tabular}{p{0.01\linewidth}p{0.95\linewidth}}
    \multicolumn{2}{c}{
\includegraphics[width=\linewidth]{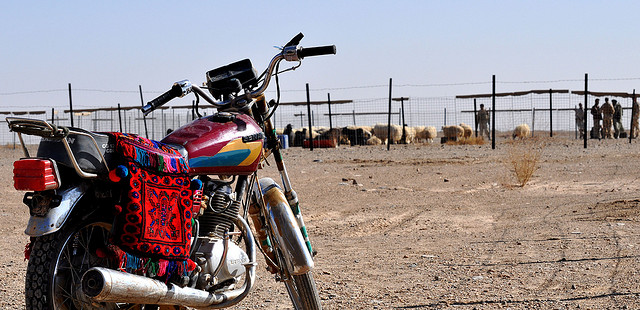}}\\ %ix21
7 &  a motorcycle parked on a dirt road\\ 
10 &  a motorcycle is parked on the side of a road\\ 
16 &  a motorcycle parked on the side of a dirt road with a fence in the background\\ 
22 &  a motorcycle parked on the side of a dirt road in front of a fence with a group of sheep behind it\\ 
28 &  a motorcycle is parked in a dirt field with a lot of sheep on the side of the road in front of a fence on a sunny day\\ 
\hline
7 & a motorcycle parked on a dirt road\\ 
10 & a motorcycle parked on a dirt road near a fence\\ 
16 & a motorcycle parked on a dirt road in front of a group of people behind it\\ 
22 & a motorcycle parked on a dirt road in front of a group of people on a dirt road next to a fence\\ 
28 & a motorcycle parked on a dirt road in front of a group of people on a dirt road in front of a group of people in the background\\ 
    \end{tabular}
%     \begin{tabular}{p{0.01\linewidth}p{0.95\linewidth}}
%     \multicolumn{2}{c}{
% \includegraphics[width=\linewidth]{figures/000000348881.jpg}}\\ %ix=18
%  7 &  an airplane is parked at an airport\\ 
% 10 &  an airplane is parked on the tarmac at an airport\\ 
% 16 &  an airplane is parked on a runway with a man standing on the side of it\\ 
% 22 &  an airplane is parked on a runway with a man standing on the side of it and a person in the background\\ 
% 28 &  an airplane is parked on the tarmac at an airport with a man standing on the side of the stairs and a man standing next to the plane\\
% \hline
%  7 & a plane is sitting on the tarmac\\ 
% 10 & a plane is sitting on the tarmac at an airport\\ 
% 16 & a plane that is sitting on the tarmac at an airport with people in the background\\ 
% 22 & a plane is sitting on the tarmac at an airport with people in the background and a man standing in the background\\ 
% 28 & a plane is sitting on the tarmac at an airport with people in the background and a man standing on the side of the road in the background\\
%     \end{tabular}
    \begin{tabular}{p{0.01\linewidth}p{0.95\linewidth}}
    \multicolumn{2}{c}{
\includegraphics[width=\linewidth]{figures/000000348881.jpg}}\\ %ix=18
 7 &  an airplane is parked at an airport\\ 
10 &  an airplane is parked on the tarmac at an airport\\ 
16 &  an airplane is parked on a runway with a man standing on the side of it\\ 
22 &  an airplane is parked on a runway with a man standing on the side of it and a person in the background\\ 
28 &  an airplane is parked on the tarmac at an airport with a man standing on the side of the stairs and a man standing next to the plane\\
\hline
 7 & a plane is sitting on the tarmac\\ 
10 & a plane is sitting on the tarmac at an airport\\ 
16 & a plane that is sitting on the tarmac at an airport with people in the background\\ 
22 & a plane is sitting on the tarmac at an airport with people in the background and a man standing in the background\\ 
28 & a plane is sitting on the tarmac at an airport with people in the background and a man standing on the side of the road in the background\\
    \end{tabular}
}
    \caption{\small Generated captions of different lengths. Top: LenEmb; Bottom: Att2in+BS10}
    \label{fig:qualitative}
\end{figure}

\noindent\textbf{Quatlitative results(Fig. \ref{fig:qualitative})} show that
the LenEmb model, when generating longer captions, changes the caption
structure and covers more detail, while the base model tends to have
the same prefix for different lengths and repetition. More results can be browsed onine~\cite{colab}.

\subsection{Failure on CIDEr optimization}
We apply SCST\cite{rennie2017self} training for length models. However, SCST doesn't work well. While the CIDEr scores can be improved, the generated captions tend to be less fluent, including bad endings (ending with 'with a') or repeating words (like 'a a').

% The length predictor will collapse to a one-hot predictor, predicting 9 words everytime. We also try fix the length predictor and train the rest of the part with RL.

% By looking at Kukichi, they seems to use the same test set, and generate different lengths, and then run the scores.

% \subsection{How attention map change given different constraints}

% \subsection{Length control comparison with SC models}

% To compare:

% golden performance table

% different length table.
% (Presumablly uniform would perform better on this table)

% \subsection{Block trigrams}
% \rtcom{todo}

\section{Conclusions}
We present two captioning models that can control the length and shows their effectiveness to generate good captions of different lengths. The code will be released at link\footnote{\url{https://github.com/ruotianluo/self-critical.pytorch/tree/length_goal}}.

{\small
\bibliographystyle{ieee_fullname}
\bibliography{egbib}
}

\end{document}